\documentclass[conference]{IEEEtran}
\IEEEoverridecommandlockouts

	\usepackage[utf8]{inputenc} 
	\usepackage[T1]{fontenc}

\usepackage{cite}

\usepackage[hidelinks]{hyperref}
\hypersetup{
    colorlinks=true,
    urlcolor=blue,
    citecolor=blue,
    unicode=false,
    pdftoolbar=true,
    pdfmenubar=true,
    pdffitwindow=false,
    pdfstartview={FitH},
    pdftitle={Extreme Learning Machines for Fast Training of Click-Through Rate Prediction Models},
    pdfsubject={Extreme Learning Machines for Fast Training of Click-Through Rate Prediction Models},
    pdfcreator={Creator},
    pdfproducer={Producer},
    pdfkeywords={residual connections} {neural networks} {CTR},
    pdfnewwindow=true,
}

\ifCLASSINFOpdf
  \usepackage[pdftex]{graphicx}
\else
\fi
\usepackage[cmex10]{amsmath}

\usepackage{multirow}
\usepackage{array}
\usepackage[lofdepth,lotdepth]{subfig}
\usepackage{rotating}

\usepackage{amsmath,amssymb,amsfonts}
\usepackage{amsthm}
\usepackage{graphicx}
\usepackage{textcomp}
\usepackage{xcolor}

\usepackage{algorithm}
\usepackage{algorithmicx}%
\usepackage{algpseudocode}
\usepackage{listings}
\usepackage[english]{babel}
\usepackage{pict2e,tikz-cd}
\usepackage{enumerate}

\newtheorem{theorem}{Theorem}
\newcommand{\argmax}{\operatornamewithlimits{arg\,max}}
\newcommand{\argmin}{\operatornamewithlimits{arg\,min}}

\begin{document}


%
\title{Extreme Learning Machines for Fast Training of Click-Through Rate Prediction Models}

\author{\IEEEauthorblockN{Ergun Biçici}
\IEEEauthorblockA{\textit{AI Enablement} \\
\textit{Huawei Türkiye R\&D Center}\\
Istanbul, Turkey \\
ergun.bicici@huawei.com}
}


%


\maketitle

\begin{abstract}

Extreme Learning Machines (ELM) provide a fast alternative to traditional gradient-based learning in neural networks, offering rapid training and robust generalization capabilities. Its theoretical basis shows its universal approximation capability. 
We explore the application of ELMs for the task of Click-Through Rate (CTR) prediction, which is largely unexplored by ELMs due to the high dimensionality of the problem. We introduce an ELM-based model enhanced with embedding layers to improve the performance on CTR tasks, which is a novel addition to the field. Experimental results on benchmark datasets, including Avazu and Criteo, demonstrate that our proposed ELM with embeddings achieves competitive F1 results while significantly reducing training time compared to state-of-the-art models such as Masknet. Our findings show that ELMs can be useful for CTR prediction, especially when fast training is needed.
\end{abstract}


\IEEEpeerreviewmaketitle

\IEEEpubidadjcol

\section{Introduction}

Gradient based learning of weights in neural networks can cause: (i) slow convergence when the learning rate is small, (ii) divergence when the learning rate is large, (iii) getting stuck at local minima, (iv) overfitting to the training data, and (v) consumption of time.
Extreme learning machine (ELM) provides an alternative to gradient based learning of single layer feedforward networks (SLFN) with fast training, good generalization, and universal approximation capability~\cite{ELMMLP2016}.
ELM is a ridge regression model with a single hidden layer whose weights and biases are randomly initialized.
ELM's performance in tasks like digit recognition, face recognition, and human action recognition is excellent in terms of accuracy/speed~\cite{ELMMLP2016}.

We define single layer feedforward neural networks (SLFNs) as follows. 
Given N distinct training instances ${(\textbf{x}_i, \textbf{y}_i) \; | \; \textbf{x}_i \in \mathcal{R}^n, \; \textbf{y}_i \in \mathcal{R}^m, \; i = 1, \ldots, N}$, an SLFN with $L$ hidden nodes can be written as~\cite{HuangZS06,tfelm,ELMMLP2016}:
\begin{equation}
    \hat{\textbf{y}}_i = \sum_{j=1}^L \beta_j h_j(\textbf{x}_i, \tilde{\textbf{w}}_j, b_j)
\end{equation}
where $h_j$ denotes the $j$th hidden node's activation function, $\textbf{w}_j$ is the input weight vector, $b_j$ is the bias of the $j$th hidden layer, and $\beta_j$ is the output weight. For activation function $g$~\cite{ELMMLP2016}:
\begin{equation}
h_j(\textbf{x}_i, \tilde{\textbf{w}}_j, b_j) = \begin{cases} g(\tilde{\textbf{w}}_j \textbf{x}_i + b_j)    & \mbox{additive} \\
g(b_j \parallel \textbf{x}_i - \tilde{\textbf{w}}_j \parallel)    & \mbox{radial basis function} \\
\end{cases}
\end{equation}

SLFNs' universal approximation capability is based on the following theorems~\cite{HuangZS06,tfelm,ELMMLP2016}:

\begin{theorem}
\label{Theorem1}
Given any bounded nonconstant piecewise continuous function $g: \mathbb{R} \rightarrow \mathbb{R}$, if span of $h(\textbf{x}, \tilde{\textbf{w}}, b): (\tilde{\textbf{w}}, b) \in \mathbb{R}^n \times \mathbb{R}$ is dense in $L^2$ space, for any target function $f$ and any function sequence $h_{1 \ldots L}(\textbf{x}, \tilde{\textbf{w}}_{1 \ldots L}, b_{1 \ldots L})$ randomly generated based on any continuous sampling distribution, $\lim_{o \rightarrow \infty} \parallel \! f - f_o \! \parallel \; = 0$ holds with probability one if the output weights are determined by ordinary least squares to minimize $\parallel \! f(\textbf{x}) - \sum_{i=1}^L \beta_i h_i(\textbf{x}, \tilde{\textbf{w}}_i, b_i) \! \parallel$.
\end{theorem}

\autoref{Theorem1} shows that networks with randomly generated weights whose output weight matrix is obtained with a least squares solution have universal approximation capability when the activation function $h$ is nonconstant piecewise and its span is dense in $L^2$ space where the $L^2$ space is a space of functions for which the $L_2$ norm, denoted as $\parallel \cdot \parallel$, is finite. 

\begin{theorem}
\label{Theorem2}
Given an SLFN with activation function $g: R \rightarrow R$ that is infinitely differentiable in any interval, for N arbitrary distinct instances $(\textbf{x}_i, \textbf{y}_i)$, $\textbf{x}_i \in \mathcal{R}^n, \; \textbf{y}_i \in \mathcal{R}^m$, for any $\tilde{\textbf{w}}_i$ and $b_i$ randomly chosen from any intervals of $\mathcal{R}^n$ and $\mathcal{R}$ respectively, according to any continuous probability distribution, then with probability one, \textbf{H} of the SLFN is invertible and $\parallel \textbf{H} \beta - \textbf{Y} \parallel = 0$.
\end{theorem}

where
\begin{equation}
\textbf{H} = \left[ \begin{array}{ccc}
h(\textbf{x}_1, \tilde{\textbf{w}}_1, b_1) & \ldots & h(\textbf{x}_1, \tilde{\textbf{w}}_L, b_L) \\
\ldots & & \ldots \\
h(\textbf{x}_N, \tilde{\textbf{w}}_1, b_1) & \ldots & h(\textbf{x}_N, \tilde{\textbf{w}}_L, b_L) \\
\end{array} \right]
\end{equation}

\section{Extreme Learning Machine}

Extreme learning machine (ELM) is a neural network model with a single hidden layer whose weights and biases are randomly initialized and the prediction is a linear 
combination of the hidden nodes where the final layer weights are learned with ridge regression~\cite{HuangZS06,ELMMLP2016}:
\begin{equation}
\begin{array}{@{\hspace{0.0cm}}lll@{\hspace{0.0cm}}}
f_{\text{ELM}}(\textbf{x}, L) & = & \displaystyle \left(\sum_{i=1}^L h(\textbf{x}, \tilde{\textbf{w}}_i, b_i) \textbf{w}_i \right) = \hat{\textbf{y}} \\
h(\textbf{x}, \tilde{\textbf{w}}_i, b_i) & = & h(\textbf{X} \tilde{\textbf{W}} + \textbf{b}) \;\;\;\;\; \text{for additive nodes} \\
& = & h(\parallel \textbf{X} - \tilde{\textbf{W}} \parallel \textbf{b}) \;\;\;\;\; \text{for RBF nodes} \\
f_{\text{ELM}}(\textbf{X}, L) & = & \displaystyle h(\textbf{X}, \tilde{\textbf{W}}, \textbf{b}) \textbf{W} = \hat{\textbf{Y}}
\end{array}
\end{equation}
where $\textbf{x} \in \mathbf{R}^{N_X \times 1}$ is an input instance and we have $N$ instances stored in $\textbf{X} \in \mathbf{R}^{N \times N_X}$, $h$ is the activation 
function, $\tilde{\textbf{W}} \in \mathbf{R}^{N_X \times L}$ is the random input weight matrix and $\textbf{b} \in \mathbf{R}^{L \times 1}$ is the random input bias, 
$\textbf{H} \in \mathbf{R}^{N \times L}$ is the hidden layer obtained, $\textbf{W} \in \mathbf{R}^{L \times N_Y}$ is the random output weight matrix, and $\hat{\textbf{y}} \in 
\mathbf{R}^{N_Y \times 1}$ is the predicted output. ELM is parameterized with the number of hidden nodes and the activation function. 

ELM learning converges when $h$ is nonconstant piecewise and its span is dense~\cite{ELMMLP2016} where the inner products $\textbf{w} \textbf{x}_1, \ldots, \textbf{w} 
\textbf{x}_n$ are different from each other with probability 1~\cite{Huang03} so that the weight matrix is guaranteed to create an N dimensional space that maps the inputs to the hidden layer for $g$ that is infinitely differentiable including sigmoid, the radial basis, sine, cosine, exponential, and many other nonregular functions~\cite{ELMActivationFunctions,HuangZS06}. 

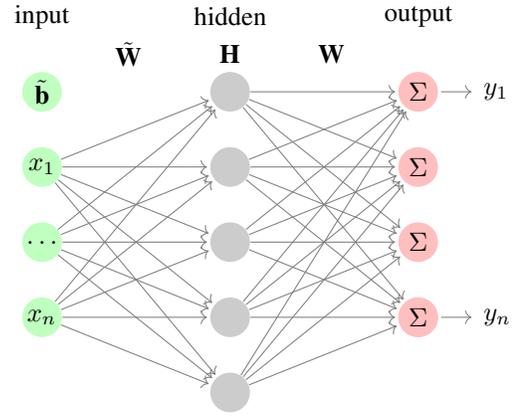
\begin{figure}[t]
\begin{center}
\begin{tikzpicture}[shorten >=1pt,->,draw=black!50, node distance=\layersep]
\def\layersep{2.5cm}
\def\layerseps{5cm}
\def\ni{3}
\def\no{4}
\def\nh{5}
\pgfmathsetmacro\inputoffsety{int(round((\nh-\ni)/2))}
\pgfmathsetmacro\middley{int(round((\nh)/2))}
    \tikzstyle{every pin edge}=[<-,shorten <=1pt]
    \tikzstyle{neuron}=[circle,fill=black!35,minimum size=15pt,inner sep=0pt]
    \tikzstyle{input neuron}=[neuron, fill=green!25];
    \tikzstyle{output neuron}=[neuron, fill=red!25];
    \tikzstyle{hidden neuron}=[neuron, fill=black!20];
    \tikzstyle{annot} = [text width=4em, text centered]

    \node[input neuron] (I) at (0,-\inputoffsety) {$\tilde{\textbf{b}}$};
    \foreach \name / \y in {1,...,1}
        \node[input neuron] (I-\name) at (0,-\inputoffsety-\y) {$x_\y$};
    \foreach \name / \y in {2,...,2}
      \node[input neuron] (I-\name) at (0,-\inputoffsety-\y) {$\ldots$};
    \foreach \name / \y in {3,...,\ni}
      \node[input neuron] (I-\name) at (0,-\inputoffsety-\y) {$x_n$};

    \foreach \name / \y in {1,...,\nh}
        \path
            node[hidden neuron] (H-\name) at (\layersep,-\y cm) {};

    \foreach \name / \y in {1,...,1}
        \node[output neuron, pin={[pin edge={->}]right:$y_\y$}] (O-\name) at (\layerseps,-\y) {$\Sigma$};
    \foreach \name / \y in {2,...,\no}
	\node[output neuron] (O-\name) at (\layerseps,-\y) {$\Sigma$};
    \node[output neuron, pin={[pin edge={->}]right:$y_n$}] (O-\no) at (\layerseps,-\no) {$\Sigma$};

    \foreach \source in {1,...,\ni}
        \foreach \dest in {1,...,\nh}
            \path (I-\source) edge (H-\dest) ;

    \foreach \source in {1,...,\nh}
        \foreach \dest in {1,...,\no}
            \path (H-\source) edge (O-\dest) ;

    \node[annot,above of=H-1, node distance=1cm] (hl) {hidden};
    \node[annot,above of=H-1, node distance=0.5cm] (HH) {$\textbf{H}$};
    \node[annot,left of=hl] {input};
    \node[annot,left of=HH, node distance=1.35cm] (W1) {$\tilde{\textbf{W}}$};
    \node[annot,right of=HH, node distance=1.35cm] (W2) {$\textbf{W}$};
    \node[annot,right of=hl] {output};
\end{tikzpicture}
\end{center}
\caption{ELM: $\tilde{\textbf{W}} \in \mathbf{R}^{N_X \times L}$ and $\tilde{\textbf{b}} \in \mathbf{R}^{N \times 1}$ 
are random and $\textbf{W} \in \mathbf{R}^{L \times N_Y}$ is learned.
Extreme learning machine contains a single hidden layer.}
\label{ELMFigure}
\end{figure}

ELM's input weights and biasses connecting the input to the hidden layer are randomly initialized and remain fixed at all times. \autoref{ELMFigure} depicts the ELM.

Instead of the gradient based learning of weights in single layer feedforward networks (SLFN), which can cause slow convergence when the learning rate is small and divergence 
when it is large, find local minima, overfit training data, and consume time~\cite{HuangZS06}, $\tilde{\textbf{W}}$ are random and $\textbf{W}$ is the OLS solution.
Some activation functions for ELM:\footnote{\url{https://en.wikipedia.org/wiki/Extreme_learning_machine}}
\begin{equation}
\begin{array}{@{\hspace{0.0cm}}llll@{\hspace{0.0cm}}}
h(\textbf{w}, \textbf{x}, b) & = & \frac{1}{1 + e^{-(\textbf{w} \textbf{x} + b)}} & \text{sigmoid} \\
 & = & \sin (\textbf{w} \textbf{x} + b) & \text{Fourier} \\
 & = & e^{b \parallel \textbf{w} - \textbf{x} \parallel} & \text{Gaussian} 
\end{array}
\end{equation}

ELM uses a linear model to learn $\textbf{W}$:
\begin{equation}
\begin{array}{@{\hspace{0.0cm}}lll@{\hspace{0.0cm}}}
\textbf{H} \textbf{W} & = & \textbf{y} \\
\textbf{W} & = & \mathrm{inv}(\textbf{H}) \textbf{y} \\
\textbf{W} & = & \mathrm{inv}_\lambda(\textbf{H}) \textbf{y} \\
\mathrm{inv}(\textbf{x}) & = & (\textbf{x}^T \textbf{x})^{-1} \textbf{x}^T \\
\mathrm{inv}_\lambda(\textbf{x}) & = & (\textbf{x}^T \textbf{x} + \lambda I)^{-1} \textbf{x}^T \\
\textbf{W} & = & (\textbf{H}^T \textbf{H} + \lambda I)^{-1} \textbf{H}^T \textbf{y} 
\end{array}
\end{equation}

The training proceeds as follows: 
\begin{enumerate}[i]
\item randomly initialize the weights, 
\item calculate the hidden layer output matrix $\textbf{H}$, 
\item obtain the output weight vector, $\textbf{W}$. 
\end{enumerate}

The ELM aims to reduce the training error like other learning models but unlike others, it also minimizes the norm of the output weights and finds such weights that are unique~\cite{ELMMLP2016,tfelm}:
\begin{equation}
minimize \;\; \parallel \textbf{W} \parallel \;\; and \;\; \parallel \textbf{H} \textbf{W} - \textbf{Y} \parallel^2 
\end{equation}
An ELM with a SLFN can be seen as a nonlinear random projection followed by ridge regression~\cite{tfelm}.

It is found that if a large neural network is used and a network with small weights with small squared error is found, then the magnitude of the weights rather than the number of weights is more important for the generalization performance~\cite{Bartlett1998}. Based on this observation, ELM should have better generalization performance than gradient based learning algorithms since they only minimize the training error~\cite{Huang2004}.

\subsection{Multilayer ELM}

Multilayer ELM (ML-ELM)~\cite{ELMAE2013}, uses an ELM autoencoder (ELM-AE) to derive features between every 
intermediate layer (\autoref{ELMAE}). An autoencoder learns latent features that reconstruct the input and reduce the noise. ELM-AE can assume that $\tilde{\textbf{W}}$, $\tilde{\textbf{b}}$, and $\textbf{W}$ are orthogonal:
\begin{equation}
\begin{array}{@{\hspace{0.0cm}}llll@{\hspace{0.0cm}}}
\textbf{H} & = & h(\tilde{\textbf{W}}, \textbf{X}, \tilde{\textbf{b}}) & \text{s.t.} \; \tilde{\textbf{W}}^T 
\tilde{\textbf{W}} = \textbf{I} \;\; \tilde{\textbf{b}}^T \tilde{\textbf{b}} = 1 \\
\textbf{W} & = & \mathrm{inv}_\lambda (\textbf{H}) \textbf{X} & \text{s.t.} \; \textbf{W}^T \textbf{W} = \textbf{I}
\end{array}
\end{equation}
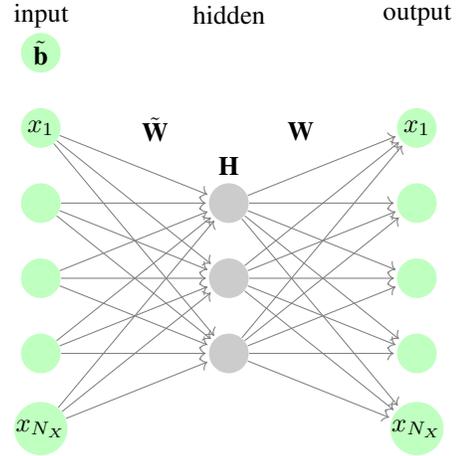
\begin{figure}[t]
\begin{center}
\begin{tikzpicture}[shorten >=1pt,->,draw=black!50, node distance=\layersep]
\def\layersep{2.5cm}
\def\layerseps{5cm}
\def\layersepss{7.5cm}
\def\ni{5}
\def\no{5}
\def\nh{3}
\pgfmathsetmacro\nimone{\ni-1}
\pgfmathsetmacro\nomone{\no-1}
\pgfmathsetmacro\inputoffsety{int(round((\ni-\nh)/2))}
\pgfmathsetmacro\middley{int(round((\nh)/2))}
    \tikzstyle{every pin edge}=[<-,shorten <=1pt]
    \tikzstyle{neuron}=[circle,fill=black!35,minimum size=15pt,inner sep=0pt]
    \tikzstyle{input neuron}=[neuron, fill=green!25];
    \tikzstyle{output neuron}=[neuron, fill=green!25];
    \tikzstyle{hidden neuron}=[neuron, fill=black!20];
    \tikzstyle{annot} = [text width=4em, text centered]
    
    \node[input neuron] (I) at (0,0) {$\tilde{\textbf{b}}$};
    \foreach \name / \y in {1,...,1}
        \node[input neuron] (I-\name) at (0,-\y) {$x_1$};
    \foreach \name / \y in {2,...,\ni}
	\node[input neuron] (I-\name) at (0,-\y) {};
    \node[input neuron] (I-\ni) at (0,-\ni) {$x_{N_X}$};
    
    \foreach \name / \y in {1,...,\nh}
        \path
            node[hidden neuron] (H-\name) at (\layersep,-\inputoffsety-\y) {};

    \foreach \name / \y in {1,...,1}
        \node[output neuron] (O-\name) at (\layerseps,-\y) {$x_1$};
    \foreach \name / \y in {2,...,\no}
	\node[output neuron] (O-\name) at (\layerseps,-\y) {};
    \node[output neuron] (O-\no) at (\layerseps,-\no) {$x_{N_X}$};

    \foreach \source in {1,...,\ni}
        \foreach \dest in {1,...,\nh}
            \path (I-\source) edge (H-\dest) ;
    
    \foreach \source in {1,...,\nh}
        \foreach \dest in {1,...,\no}
            \path (H-\source) edge (O-\dest) ;
    
    \node[annot,above of=H-1, node distance=2.5cm] (hl) {hidden};
    \node[annot,above of=I, node distance=0.5cm] {input};
    \node[annot,above of=O-1, node distance=1.5cm] {output};
    \node[annot,above of=H-1, node distance=0.5cm] {$\textbf{H}$};
    \node[annot,above left of=H-1, node distance=1.39cm] (W1) {$\tilde{\textbf{W}}$};
    \node[annot,above right of=H-1, node distance=1.35cm] (W2) {$\textbf{W}$};
\end{tikzpicture}
\end{center}
\caption{ELM-AE contain a single hidden layer.}
\label{ELMAE}
\end{figure}
Orthogonal procrustes solves orthogonal ridge regression:
\footnote{\url{https://jianfengwang.files.wordpress.com/2015/07/proof_of_the_orthogonal_procrustes_problem.pdf}}
\begin{equation}
\begin{array}{@{\hspace{0.0cm}}lll@{\hspace{0.0cm}}}
\textbf{W} & = & \displaystyle \argmin_{\textbf{W}^\prime} \; \parallel \textbf{X} \textbf{W}^\prime - \textbf{Y} 
\parallel \;\; \text{s.t. } {\textbf{W}^\prime}^T \textbf{W}^\prime = \textbf{I} \\
& = & \displaystyle \argmin_{\textbf{W}^\prime} \; \mathrm{tr}((\textbf{X} \textbf{W}^\prime - \textbf{Y})^T 
(\textbf{X} \textbf{W}^\prime - \textbf{Y})) \\
& = & \displaystyle \argmin_{\textbf{W}^\prime} \; \mathrm{tr}(\textbf{Y}^T \textbf{Y}) + \mathrm{tr}(\textbf{X}^T 
\textbf{X}) - 2 \mathrm{tr}(\textbf{Y}^T \textbf{X} \textbf{W}^\prime) \\
& = & \displaystyle \argmax_{\textbf{W}^\prime} \; \mathrm{tr}(\textbf{Y}^T \textbf{X} \textbf{W}^\prime) \\
& = & \displaystyle \argmax_{\textbf{W}^\prime} \; \mathrm{tr}(\textbf{W}^\prime \textbf{Y}^T \textbf{X}) \;\;\;\;\; 
\mathrm{tr}(\textbf{A}\textbf{B}) = \mathrm{tr}(\textbf{B}\textbf{A}) \\
& = & \displaystyle \argmax_{\textbf{W}^\prime} \; \mathrm{tr}(\textbf{W}^\prime \textbf{U} \pmb{\Sigma} \textbf{V}^T) 
\;\;\;\;\; \textbf{U} \pmb{\Sigma} \textbf{V}^T = \mathrm{svd}(\textbf{Y}^T \textbf{X}) \\
& = & \displaystyle \argmax_{\textbf{W}^\prime} \; \mathrm{tr}(\textbf{V}^T \textbf{W}^\prime \textbf{U} \pmb{\Sigma}) 
\\
& = & \displaystyle \argmax_{\textbf{W}^\prime} \; \mathrm{tr}(\hat{\textbf{W}^\prime} \pmb{\Sigma}) \;\;\;\;\; 
\hat{\textbf{W}^\prime} = \textbf{V}^T \textbf{W}^\prime \textbf{U}
\\
& = & \textbf{V} \textbf{U}^T
\end{array}
\end{equation}
Thus, $\mathrm{svd}(\textbf{X}^T \textbf{H}) = \textbf{U} \pmb{\Sigma} \textbf{V}^T$.
For ridge regression, we use $\mathrm{svd}(\mathrm{inv}_\lambda(\textbf{H}) \textbf{X}) = 
\mathrm{svd}(\textbf{W}_{\text{ridge}})$.

The first step in ML-ELM is ELM-AE which obtains W. The transpose of this W is used as the first layer in ML-ELM~\cite{tfelm,ELMTrends2013}. Then learning proceeds as a regular ELM model.

\section{Experimental Results}
\label{Experiments}

Click-through rate (CTR) of advertisements (ads) represents the percentage of ads that are actually clicked by users. CTR prediction models predict whether the user will click on an ad or not after being shown the ad. 
Previous work on using ELMs for CTR prediction involves only small scale experiments~\cite{HierarchicalELMCTR2018}. They use a hierarchical two-level ELM model where the second level ELM model uses the predictions of the first level ELM models that are trained on different feature subsets of the same dataset. The experiments were done on a dataset with only four features on 5 different datasets with the number of instances ranging from 4440 to 318455. The largest dataset they use is 1400 times smaller than the Criteo dataset even if we accept all features as equal.


\subsection{Datasets and Implementation Details}

The training and test datasets for CTR are unbalanced since the impressions that actually result with clicks can be as low as $2\%$ to $10\%$. We predict CTR results using the publicly available Avazu dataset~\cite{Avazu}. We use the 8/1/1 split for training/validation/test sets~\cite{AvazuX4,DBLP:conf/sigir/ZhuDSMLCXZ22}.
Avazu dataset contains 23 features about the user and the device that is used to connect to the website. The features include the hour, website domain and category, device model and type, and advertisement banner position. The sizes of the datasets are 32,343,172, 4,042,897, and 4,042,898 instances correspondingly for training, validation, and test sets. Its CTR is $0.17$.

Additionally, we use the publicly available Criteo dataset with 8/1/1 split for training/validation/test sets~\cite{CriteoX4}.
Criteo dataset contains 39 features about which 13 of them are integer features and 26 of them are categorical and some features can have missing values. The sizes of the datasets are 36,672,493, 4,584,062, and 4,584,062 instances correspondingly for training, validation, and test sets. Its CTR is $0.256$. We also use a company dataset for CTR prediction. The dataset contains about 700 million instances and its CTR is about $0.08$. 

In our models, the batch size is set to 10000. 
We use a hidden layer with 1000 neurons. 
We use the ReLU activation function. 
We compare the ELM results with the Masknet~\cite{Masknet} model. 
The embedding dimension for the Masknet model is set to 8. 
For Masknet, we initialize the learning rate to 7e-3 and automatically reduce the learning rate when a plateau is reached.
We repeat each experiment for 3 times and present averaged results to decrease errors. Out implementation is based on the tfelm package.~\footnote{\url{https://github.com/popcornell/tfelm}.}

\subsection{Evaluation Metrics}

The evaluation metrics we use are AUC (area under ROC, receiver operating characteristic, curve), loss or the loss value we obtain from our loss functions, precision of class $1$ labels or clicks, corresponding recall, and $F_1$ or the harmonic average of precision and recall. AUC can be interpreted as the probability that the model assigns a randomly selected positive instance a higher rank than a randomly selected negative instance~\cite{AUCInterpretation}.
Except for the loss, the higher the metric the better the performance.
We also use the early stopping in Tensorflow training for Masknet. The "patience" parameter with early stopping is set to $3$. 

\subsection{ELM with Embedding Layer}

In high dimensional settings where feature values can take millions of different categorical values, embeddding layers or matrices are used. Embedding techniques can represent high-dimensional categorical data in a lower-dimensional space while preserving important relationships between categories. An embedding model, such as Word2Vec~\cite{mikolov2013efficient} or GloVe~\cite{pennington2014glove}, can be trained on a large corpus of data to learn a dense vector representation for each category. These embeddings can then be used as inputs to a neural network for machine learning tasks. 
Since ELM is not a gradient-based learning technique, the embedding layer will not be trained or updated if used within an ELM architecture. To use learned embedding matrices, we first use the DeepFM model~\cite{DeepFM}, which uses an embedding layer and train it for one epoch. Then we use the trained embedding weights to transform the input to the ELM model rather than using the original input features.

\subsection{Performance Evaluation}

\autoref{ResultsTable} presents our ELM and ML-ELM results compared with plain Masknet results on both the Avazu and Criteo datasets. ELM emb and ML-ELM emb results use an embedding layer trained with DeepFM and saved for later use with the ELM models. The results show that Masknet with gradient-learning obtains larger AUC in both Avazu and Criteo datasets. However, training the Masknet model takes several epochs and hours. In contrast, training the ELM models take minutes where the ML-ELM model takes about twice as long as the ELM due to the ELM-AE step. 

\begin{table}[t]
    \centering
  \caption{\textsc{Experimental results.}}
    \begin{tabular}{ll|lll}
    \hline
        Dataset & model & logloss & AUC & F1 \\ \hline
        Avazu & Masknet & 0.39595 & 0.77606 & 0.30103 \\
        Avazu & ELM & 0.64184	& 0.56468	& 0.14997 \\
        Avazu & ELM emb & 0.72294	& 0.68308	& \textbf{0.34791} \\
        Avazu & ML-ELM & 0.64184	& 0.56468	& 0.14997 \\ 
        Avazu & ML-ELM emb & 0.44229	& 0.71713	& 0.04839 \\ \hline
        Criteo & Masknet & 0.46144 & 0.7878 & 0.44829 \\
        Criteo & ELM & 0.56277	& 0.57076	& 0.00221 \\
        Criteo & ELM emb & 9.54421	& 0.68152	& 0 \\
        Criteo & ML-ELM & 0.56432	& 0.55906	& 0.00069 \\ 
        Criteo & ML-ELM emb & 0.67759	& 0.65275	& \textbf{0.45041} \\ \hline
    \end{tabular}
  \label{ResultsTable}
\end{table}

The results in \autoref{ResultsTable} show that the addition of the embedding layer improves the AUC significantly for both ELM and ML-ELM models. We do not know any prior work that have tried to use embedding weights with ELM models. Additionally, according to the F1 results, ELM emb model obtains better results than Masknet on the Avazu dataset while the ML-ELM emb model obtains better results than Masknet on the Criteo dataset.

\begin{table}[!ht]
    \centering
  \caption{\textsc{Results on the company dataset. \# neurons represent the number of neurons in each layer separated by a ";".}}
    \begin{tabular}{ll|llll}
    \hline
         & \# neurons & logloss & auc & f1 & batch time \\ \hline
         Masknet & 400;400 & 0.15358 & 0.93040 & \\ \hline
        \multirow{7}{*}{ELM} & 500 & 0.2381 & 0.6378 & 0.00955 & 0.07403 \\
         & 1000 & 0.2375 & 0.6443 & 0.01255 & 0.07748 \\
         & 2000 & 0.2376 & 0.6525 & 0.01705 & 0.07905 \\
         & 5000 & 0.2360 & 0.6672 & 0.02532 & 0.08508 \\
         & 10000 & 0.2369 & 0.6754 & 0.03608 & 0.08649 \\
         & 20000 & 0.2366 & 0.6855 & 0.05191 & 0.35536 \\
         & 40000 & 0.2297 & 0.6945 & 0.06280 & 1.41308 \\ \hline
        \multirow{6}{*}{ML-ELM} & 500;500 & 0.2410 & 0.6074 & 0 & 0.08802 \\
         & 1000;1000 & 0.2441 & 0.6162 & 0 & 0.08992 \\
         & 2000;2000 & 0.2462 & 0.6245 & 0.00010 & 0.09088 \\
         & 5000;5000 & 0.3215 & 0.6320 & 0.01746 & 0.09458 \\
         & 7500;7500 & 0.5417 & 0.6061 & 0.04379 & 0.09895 \\
         & 10000;10000 & 0.2590 & 0.6189 & 0.01652 & 0.16841 \\ \hline
    \end{tabular}
  \label{HuaweiDatasetResults}
\end{table}

In~\autoref{HuaweiDatasetResults}, our results on the company dataset is listed. On this real dataset, ELM and ML-ELM falls short of obtaining close results to  Masknet. We have not tried ELM models with the embedding layer as this requires saving DeepFM weights locally and reloading them.
In~\autoref{HuaweiDatasetResults}, we tried the ELM model with increasing number of neurons, which gradually improves the performance but after 10,000 neurons, the time for processing a batch of the dataset significantly increases. Up to 10,000 neurons, the ELM model takes about 6 minutes of training while the Masknet that converges on two epochs takes about 15 minutes on this dataset.

\section{Conclusion}
\label{Conclusion}

We applied extreme learning machines to the task of click-through rate prediction, which is a very high dimensional problem. We enhanced ELM models including the basic ELM and the multi-layer ELM with embedding layers to obtain improvements in the performance. Using ELMs together with embedding layers is unexplored before. With this addition, we are able to obtain improvements in F1 results compared with a state-of-the-art CTR prediction model, Masknet. We demonstrate our results on two benchmark datasets, namely the Avazu and Criteo datasets, as well as a company dataset. We could not improve the performance in terms of AUC or logloss but we reduce the training time significantly with ELM models. 
Our findings show that ELMs can be useful for CTR prediction, especially when fast training is needed. It can be useful for rapid model training and deployment. 
ELM models without embedding layers did not perform as well as Masknet on the company dataset, therefore future work involves enhancements in feature representation.

\bibliographystyle{IEEEtran}







\end{document}